\documentclass[letterpaper,10pt,conference]{ieeeconf}

\IEEEoverridecommandlockouts
\usepackage{cite}
\usepackage{graphicx}
\usepackage{epsfig}
\usepackage{amsmath}
\usepackage{amssymb}
\usepackage{booktabs}
\usepackage{multirow}
\usepackage{url}
\usepackage{hyperref}

\usepackage{amsmath,amsfonts,bm}

\def\eqref#1{equation~\ref{#1}}
\def\1{\bm{1}}

\DeclareMathAlphabet{\mathsfit}{\encodingdefault}{\sfdefault}{m}{sl}
\SetMathAlphabet{\mathsfit}{bold}{\encodingdefault}{\sfdefault}{bx}{n}

\newcommand{\sout}[1]{}

\title{\LARGE \bf
FastGrasp: Learning-based Whole-Body Control Method for Fast Dexterous Grasping with Mobile Manipulators
}

\author{
Heng Tao$^{1*}$, Yiming Zhong$^{1*}$, Zemin Yang$^{1*}$, Yuexin Ma$^{1\dagger}$
\thanks{$^{1}$ShanghaiTech University
        {\tt\small \{taoheng2023, mayuexin\}@shanghaitech.edu.cn}}%
\thanks{$^{*}$Equal contribution }%
\thanks{$^{\dagger}$Corresponding author}%
}

\begin{document}
\maketitle
\thispagestyle{empty}
\pagestyle{empty}

\begin{abstract}
Fast dexterous grasping while a mobile base remains in motion requires coordinated whole-body control and rapid adaptation to physical contact. We propose \textbf{FastGrasp}, a two-stage learning framework that integrates grasp guidance, whole-body control, and tactile feedback. First, a pretrained conditional variational autoencoder generates diverse grasp candidates from object point clouds. Candidates are filtered by approach direction and supporting-surface constraints, then ranked using an
envelopment-based criterion combining grasp width and depth coverage measures. Second, a reinforcement learning policy jointly controls the mobile base, arm, and dexterous hand, using the selected grasp as guidance and tactile observations for online grasp adjustment. The policy is trained with domain randomization and deployed with command filtering and tactile-triggered grasp tightening. Simulation experiments show higher grasp success rates than
the evaluated baselines under full and partial point-cloud
observations, while real-world experiments demonstrate
sim-to-real transfer across diverse object geometries. Our project page is available at \href{https://taoheng-star.github.io/fastgrasp-page/}{https://taoheng-star.github.io/fastgrasp-page/}.
\end{abstract}

\section{INTRODUCTION}

Fast grasping enables dexterous mobile robots to grasp target objects while moving, supporting applications in logistics, domestic service, and industrial manufacturing. By coordinating mobility with manipulation rather than separating them into consecutive operations, robots can potentially improve sorting efficiency, service responsiveness, and production throughput. This capability extends manipulation from stationary grasp execution to interaction with objects during base motion, representing an important step toward more dynamic and capable robotic manipulation.

However, existing approaches remain limited in addressing fast mobile grasping as an integrated problem. Fixed-base dexterous manipulation methods~\cite{wei2024d,zhang2025robustdexgrasp,zhang2025adg} achieve precise grasp synthesis and execution, but are restricted to the manipulator's reachable workspace. Mobile manipulation systems~\cite{burgess2023architecture,haviland2022holistic,burgess2024reactive} extend workspace coverage through coordinated base and arm motion, yet typically rely on simple grippers with limited dexterity. Tactile-enhanced methods~\cite{chebotar2014learning,lee2024dextouch} complement visual perception with contact information, but are not specifically designed for rapid contact-dependent adaptation during high-speed grasping. Therefore, fast mobile grasping requires these capabilities to be tightly integrated, so that geometric grasp selection, whole-body motion, and contact-dependent adjustments remain coordinated throughout execution under strict temporal constraints.

This task presents three main challenges. \textbf{(i) Maintaining a stable grasp during motion:} Contact during a moving approach can disturb the object, leading to rebound, rotation, or slippage. The limited time available for establishing a grasp makes execution sensitive to timing and pose errors. \textbf{(ii) Real-time whole-body coordination:} Fast dexterous grasping requires seamless coordination of the mobile base, arm, and multi-fingered hand under high-speed motion. The robot must coordinate base motion with end-effector positioning and finger closure throughout approach, grasp, and withdrawal, rather than treating navigation and grasping as independent operations. \textbf{(iii) Grasping objects of varying shapes and sizes:} Differences in object geometry and pose affect feasible hand configurations and contact patterns, requiring adaptive grasp guidance and execution strategies that accommodate these variations.

\begin{figure}
    \centering
    \includegraphics[width=1.0\linewidth]{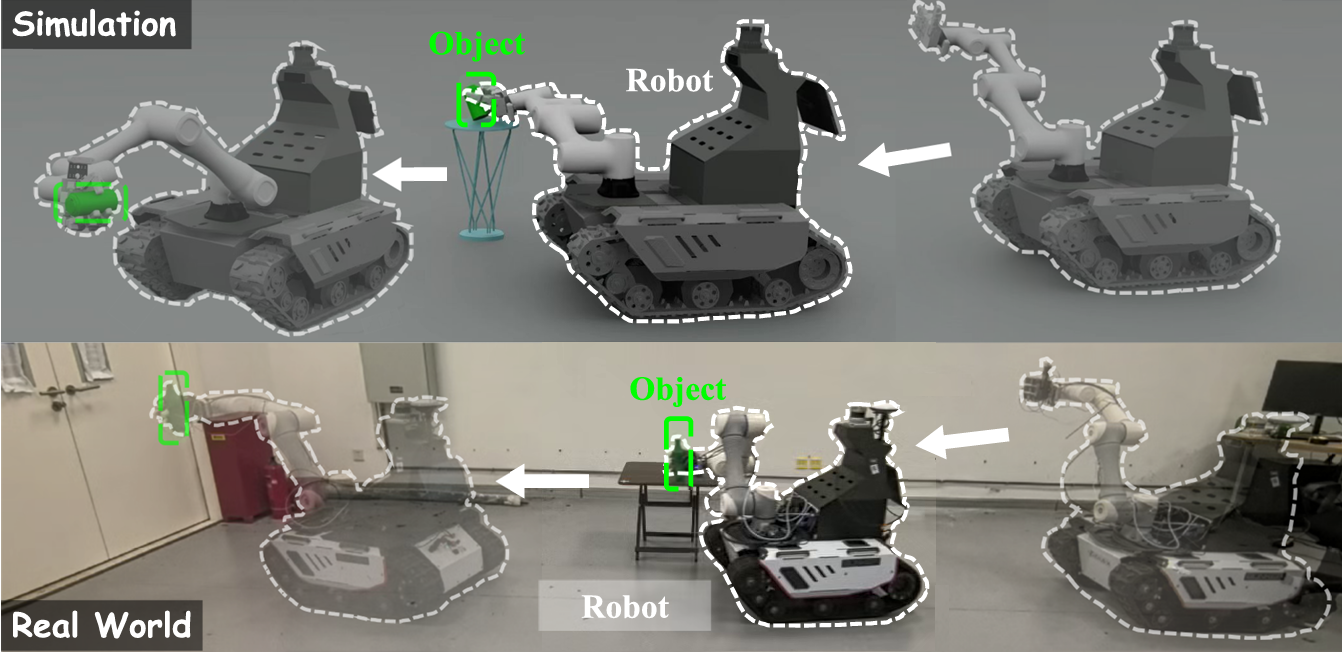}
    \caption{\textbf{FastGrasp} demonstrations in simulation and the real world.}
    \vspace{-1ex}
    \label{fig:teaser}
\end{figure}

To address these challenges, we propose \textbf{FastGrasp},
a learning-based framework that integrates grasp guidance,
whole-body control, and tactile feedback for fast mobile
grasping.
To accommodate diverse object geometries, a pretrained grasp
generator first proposes candidate grasps, which are filtered
according to approach direction and supporting-surface
constraints and ranked by hand--object envelopment.
The selected grasp then guides a reinforcement learning policy
that jointly controls the mobile base, arm, and dexterous hand,
addressing the whole-body coordination required during rapid
approach and grasping.
Tactile observations and a contact-based reward support
contact-aware policy learning, while tactile-triggered finger
tightening provides rapid adjustment during real-world
deployment.
The policy is trained in simulation and transferred to the
physical robot using domain randomization and command
filtering. Our key contributions are summarized as follows:

\begin{enumerate}

\item We propose FastGrasp, a learning-based whole-body control framework that enables coordinated control of the mobile base, arm, and dexterous hand for fast grasping while the robot remains in motion.

\item We introduce a grasp guidance selection method that filters candidates by approach direction and supporting-surface constraints, then ranks them using grasp width and depth
coverage to guide whole-body policy learning in fast grasping scenarios.

\item We develop a tactile-assisted execution strategy that
integrates tactile feedback into policy learning and
enables tactile-triggered finger tightening during
real-world deployment.

\end{enumerate}

We validate FastGrasp through simulation comparisons,
component-wise ablations, and real-world deployment across
diverse object geometries.

\section{RELATED WORK }

\subsection{Dexterous Manipulation}

Dexterous manipulation has received substantial attention in
robotics~\cite{liu2024realdex,ze20243d,wang2025dexh2r,
dasari2022learning,zhang2025adg,zhong2025dexgrasp,
zhu2025evolvinggrasp,li2023gendexgrasp,wang2022dexgraspnet,andrychowicz2020learning}.
Multi-fingered hands~\cite{shadowrobot2023dexterous}
support a range of contact configurations and manipulation
behaviors~\cite{bicchi2002hands,okamura2000overview}, but
their high-dimensional control spaces make policy learning
challenging.
Existing approaches address this problem through
goal-conditioned execution~\cite{xu2023unidexgrasp},
grasp prediction from robot and object representations
~\cite{wei2024d}, and hand-centric shape representations
~\cite{zhang2025robustdexgrasp}.
These developments provide useful tools for grasp generation
and execution. Our focus is on extending grasp guidance to
mobile systems, where objects may lie beyond the
arm's reach and grasp execution must remain coordinated with
base motion. This setting requires coupling the selected hand
configuration with the motion of the entire mobile manipulator.

\subsection{Whole-Body Control for Mobile Manipulation}

Mobile manipulation can be organized as sequential base and
arm execution~\cite{wang2009hybrid} or coordinated whole-body
control. 
Optimization-based approaches%
~\cite{shan2004motion,burgess2023architecture,
burgess2024reactive,haviland2022holistic}
use model-based objectives and constraints to coordinate
mobility and manipulation, while learning-based approaches%
~\cite{zhou2025hierarchical,zhang2024catch,fu2024mobile}
learn control policies for task execution.
In particular, Catch It!~\cite{zhang2024catch} combines a
mobile base, arm, and dexterous hand for catching objects
in flight through a two-stage training strategy for tracking
and catching. UniDexGrasp~\cite{xu2023unidexgrasp} combines
grasp proposal generation with goal-conditioned execution
for tabletop grasping.
FastGrasp instead focuses on table-supported dexterous
grasping while the mobile base remains in motion, requiring
grasp guidance to be coupled with coordinated base--arm--hand
control throughout execution.

\subsection{Tactile-Based Manipulation}

Tactile sensing provides contact information for tasks such as
in-hand object localization~\cite{chebotar2014learning},
object recognition~\cite{liu2017recent},
deformable-object manipulation~\cite{kaboli2016tactile},
and wire manipulation for assembly~\cite{pirozzi2018tactile}.
DexTouch~\cite{lee2024dextouch} demonstrates dexterous
manipulation using binary tactile feedback without visual
observations.
For fast mobile grasping, FastGrasp uses binary tactile
feedback to complement geometric grasp guidance during
coordinated base, arm, and hand motion.
Our design emphasizes timely contact detection and
contact-dependent grasp adjustment during on-the-move
grasping, rather than continuous contact-force estimation.

\section{System Setup}
\label{sec:System_Setup}
\begin{figure}[t]
    \centering
    \includegraphics[width=1\linewidth]{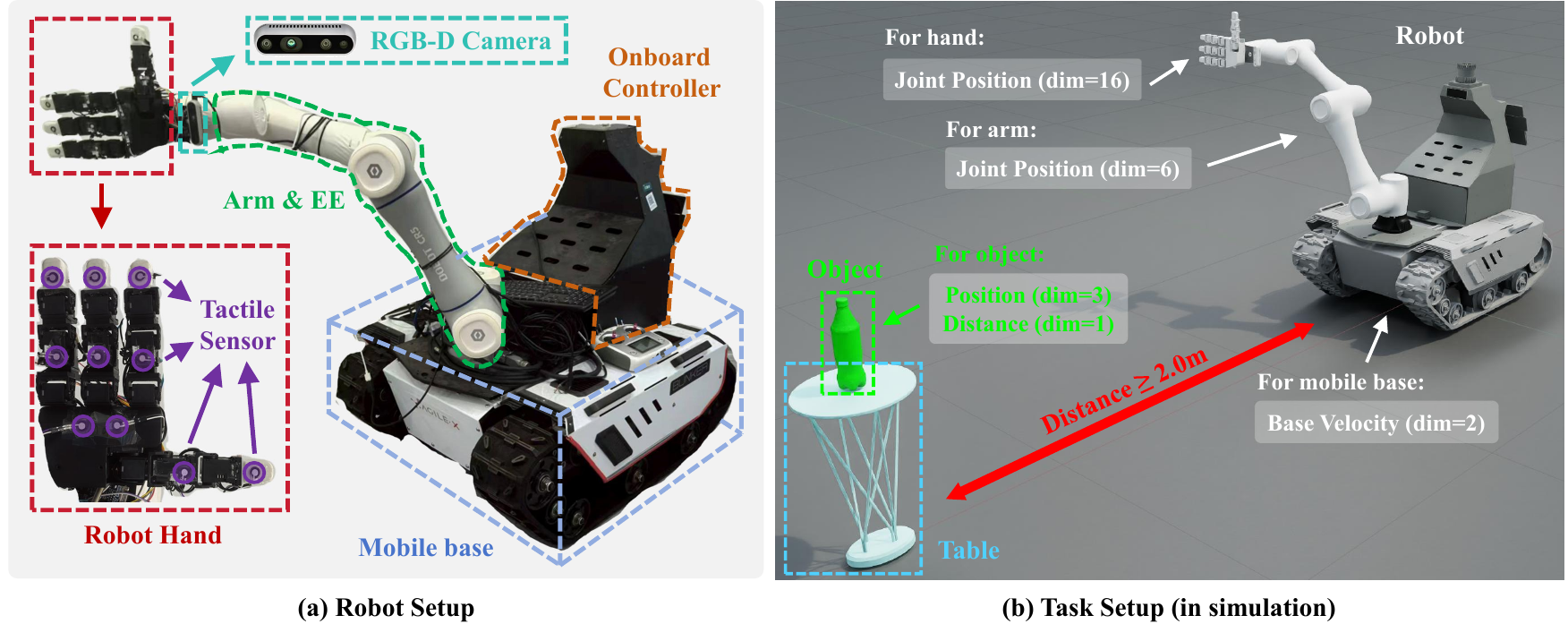}
    \caption{Real-world robot platform and simulation setup.}
    \label{fig:Environment Setup}
    \vspace{-2ex}
\end{figure}

\textbf{Real-world Setup.}
As shown in Fig.~\ref{fig:Environment Setup}, the robot
consists of an Agilex Bunker Mini mobile base, a 6-DOF
Dobot CR5 arm, and a 16-DOF LEAP Hand.
Nine resistive thin-film pressure sensors are
distributed across the fingers and palm to obtain binary
tactile information.
A RealSense D435i camera mounted at the arm's end-effector
provides egocentric RGB-D observations.
An NVIDIA Jetson AGX Orin performs onboard computation.
The RGB-D observations are used to obtain the target point
cloud, and the policy performs inference and publishes
control commands at 15 Hz, matching the simulation policy
frequency.

\textbf{Simulation Setup.}
We use Isaac Sim~\cite{gao2026nvidiaisaacsimenabling}
as the simulation environment.
As shown in Fig.~\ref{fig:Environment Setup}, the robot
starts at rest, with the target object placed on a table
at least 2.0 m ahead to provide sufficient distance for
base acceleration.
Each episode terminates and resets upon an unintended
collision or an object drop.
At reset, the robot returns to its initial state, while
the object's position within the illustrated region and
its orientation on the table are randomized.
To increase training diversity, each parallel environment
is initialized with a distinct object and a randomized
table height.

\textbf{Robot and Control Setup.}
URDF models reproduce the physical robot's nominal geometry
and kinematics in simulation.
\textbf{(i) Mobile base:}
The controller specifies forward velocity and yaw rate,
limited to 1.3 m/s and 1.0 rad/s, respectively.
The base accelerates from rest to its maximum forward-speed
command in approximately 0.7 s.
In the primary setting, this maximum command is reached before
hand--object contact and maintained during the subsequent
grasp approach.
\textbf{(ii) Arm:}
The arm is controlled through joint position commands,
with a maximum joint speed of $100^{\circ}$/s.
For safety during high-speed motion, the fourth arm joint is
fixed at the horizontal configuration used throughout our
experiments.
\textbf{(iii) Hand:}
Joint position commands control the 16-DOF hand,
with a maximum joint speed of $90^{\circ}$/s.

In this study, \emph{fast grasping} refers to dexterous
grasping performed while the mobile base remains in motion,
without stopping before grasp execution.
Our primary high-speed setting uses maximum forward-velocity
and yaw-rate commands of 1.3 m/s and 1.0 rad/s, respectively,
while an additional reduced-speed setting uses 0.65 m/s and
0.5 rad/s for comparison.
Prior work~\cite{burgess2023architecture}
reports quantitative pick-and-place results at base speeds
of 0.2--0.4 m/s and an additional grasping demonstration
at 0.8 m/s.
Maintaining base motion throughout grasp execution reduces
the available manipulation time, requiring tight coordination
of base motion, end-effector positioning, and finger closure.

\section{Learning Fast Dexterous Grasping with Tactile Feedback}

\subsection{Overview}

\begin{figure*}
    \centering
    \includegraphics[width=0.95\linewidth]{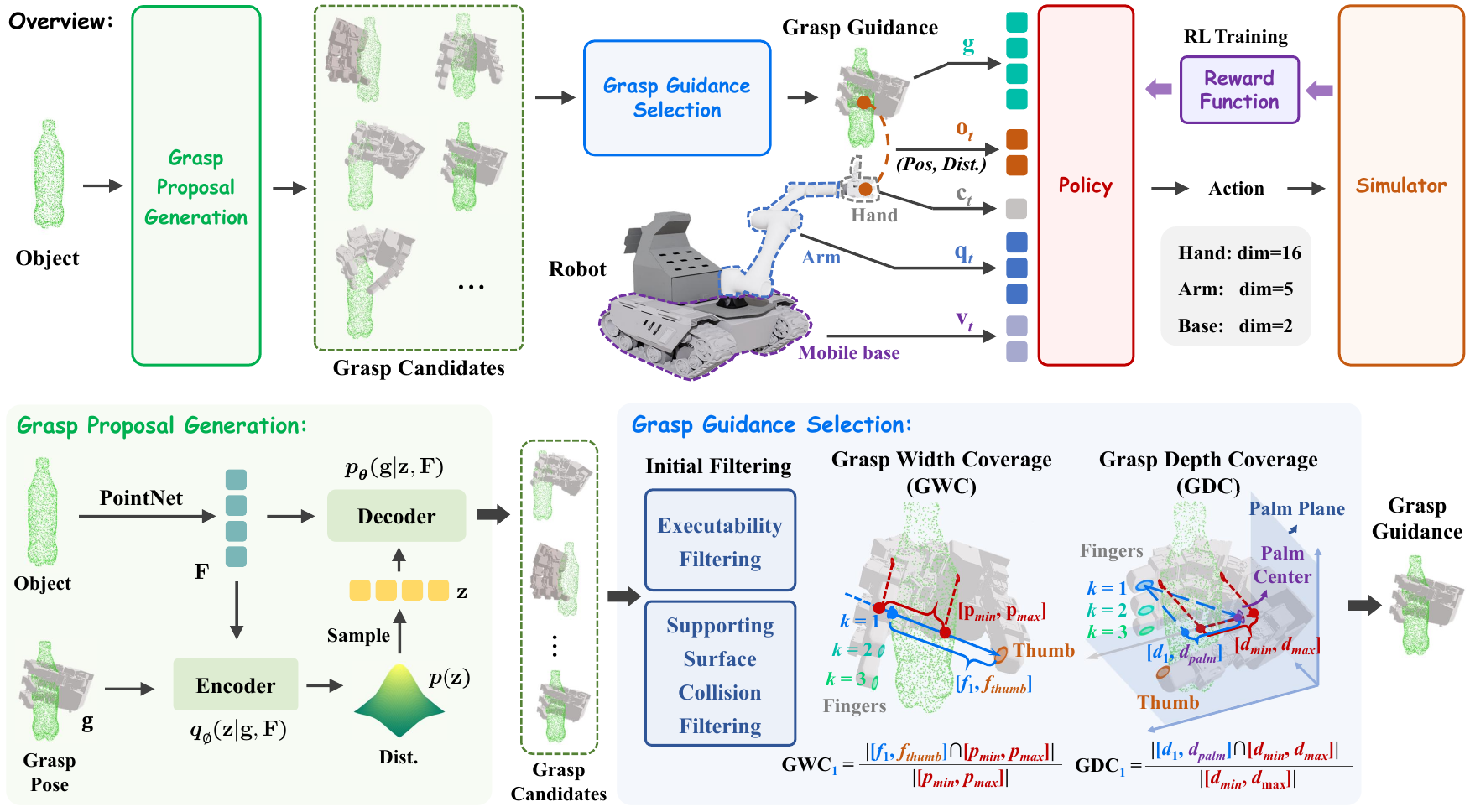}
    \caption{Overview of the FastGrasp framework.}    \label{fig:pipeline}
    \vspace{-2ex}
% 要展示cvae模型框架
% 筛选方法展示
    
\end{figure*}

Our goal is to enable a mobile manipulator to approach objects
beyond its arm reach, grasp them while the base remains
in motion, and withdraw from the table while retaining the
object. Inspired by grasp-guided execution~\cite{xu2023unidexgrasp},
FastGrasp follows a two-stage pipeline, as illustrated in
Fig.~\ref{fig:pipeline}.

\textbf{First, grasp guidance generation and selection.}
We pretrain a conditional variational autoencoder (CVAE) to
generate grasp candidates from object point clouds
(Section~\ref{sec:Grasp_Proposal_Generation}).
Candidates are filtered according to approach direction and
supporting-surface constraints and ranked by hand--object
envelopment.
The highest-scoring candidate is used as grasp
guidance for policy learning
(Section~\ref{sec:Grasp_Guidance_Selection}).

\textbf{Second, policy learning with tactile feedback.}
A reinforcement learning policy uses the selected grasp
guidance, robot and object states, and tactile
observations to generate coordinated control commands.
A contact-based reward encourages hand--object
contact during training
(Section~\ref{sec:Learning Fast Grasping Policy Using Binary Tactile Sensors}).
During real-world deployment, tactile-triggered finger
tightening complements the learned controller.

\subsection{Grasp Proposal Generation}
\label{sec:Grasp_Proposal_Generation}

Our goal is to generate diverse grasp candidates conditioned
on an object's point cloud.
Following the conditional generation approach
in~\cite{mayer2022ffhnet}, we use a CVAE~\cite{sohn2015learning}
for its ability to rapidly sample diverse grasp candidates.
PointNet~\cite{qi2017pointnet} encodes the object point cloud
into a feature vector $\mathbf{F}$.
During training, the encoder receives $\mathbf{F}$ and a
successful grasp configuration $\mathbf{g}$ and predicts an
approximate posterior
$q_{\phi}(\mathbf{z}\mid\mathbf{g},\mathbf{F})$.
The decoder reconstructs the grasp from a latent sample
drawn from this posterior and the object feature.
We maximize the evidence lower bound:

\begin{equation}
\begin{aligned}
\mathcal{L}_{\mathrm{CVAE}}(\theta,\phi;\mathbf{g},\mathbf{F})
&=\mathbb{E}_{q_{\phi}(\mathbf{z}\mid\mathbf{g},\mathbf{F})}
\left[\log p_{\theta}(\mathbf{g}\mid\mathbf{z},\mathbf{F})\right]\\
&\quad-D_{\mathrm{KL}}\!\left(
q_{\phi}(\mathbf{z}\mid\mathbf{g},\mathbf{F})\|p(\mathbf{z})\right).
\end{aligned}
\label{eq:cvae}
\end{equation}

Here, $p(\mathbf{z})=\mathcal{N}(\mathbf{0},\mathbf{I})$
is the latent prior, and
$p_{\theta}(\mathbf{g}\mid\mathbf{z},\mathbf{F})$
denotes the conditional distribution modeled by the decoder.
The reconstruction term encourages the model to reproduce
successful training grasps conditioned on the object feature,
while the KL term regularizes the approximate posterior
toward the prior.

During inference, we sample $\mathbf{z}\sim p(\mathbf{z})$
and concatenate it with $\mathbf{F}$ as input to the decoder
to generate a grasp candidate.
We generate 150 candidates for each target object to provide
diverse grasp configurations for guidance selection.
The pretrained generator remains frozen during policy training.

\subsection{Grasp Guidance Selection}
\label{sec:Grasp_Guidance_Selection}

The generator produces diverse grasps, but some candidates
may be unsuitable for execution during a moving approach.
For example, rear-side grasps may require additional
end-effector reorientation or motion around the object,
making them difficult to execute within the limited time
available for fast grasping.
We therefore filter candidates before ranking them according
to the extent of hand--object envelopment. The filtering
process considers two aspects:

\textbf{Executability filtering.}
We define a robot-facing forward-grasp reference according
to the direction from the target object toward the robot.
For each candidate, we evaluate its orientation deviation
from this reference and reject candidates whose deviation
exceeds $90^{\circ}$.
This heuristic favors robot-facing grasps and reduces
unnecessary end-effector reorientation or motion around the
object, which is particularly important under the limited
execution time of fast grasping.

\textbf{Supporting-surface collision filtering.}
To reduce potential collisions with the supporting surface,
we reject candidates whose end-effector position lies below
the object's lowest point.
This height-based criterion provides an efficient
supporting-surface collision check without requiring explicit
mesh-based collision detection.

The remaining candidates are evaluated using two complementary geometric measures: Grasp Width Coverage (GWC) and Grasp Depth
Coverage (GDC). GWC measures enclosure along the thumb--finger
directions, whereas GDC measures enclosure along the palm-normal
direction. Both measures are computed from point-cloud projections
and therefore do not require object-surface normal estimation,
which is beneficial under partial or noisy observations. Together,
they favor grasp configurations that provide greater hand--object
envelopment.

\begin{figure}
    \centering
    \includegraphics[width=0.95\linewidth]{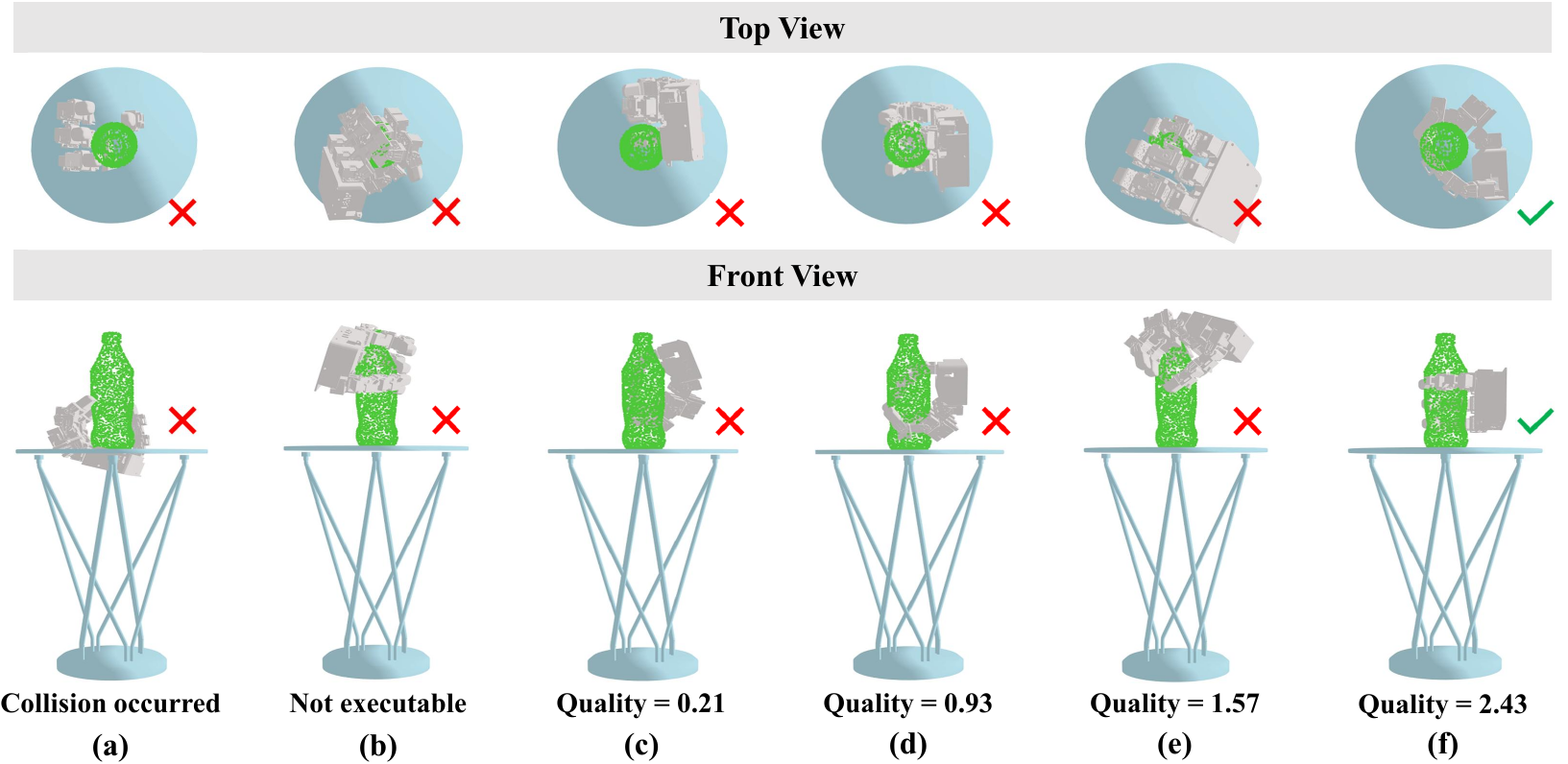}
    \caption{Grasp guidance selection.}
    \label{fig:Grasp_Guidance_Selection}
    \vspace{-2ex}
\end{figure}

\textbf{Grasp Width Coverage (GWC).}
Let $\mathcal{P}$ denote the object point cloud and $\mathbf{f}_{k}$ and $\mathbf{f}_{\mathrm{thumb}}$ the 3D positions of fingertip $k$ and the thumb tip. For $k\in\{1,2,3\}$, the thumb--finger axis has unit direction
$\mathbf{w}_{k}=(\mathbf{f}_{\mathrm{thumb}}-\mathbf{f}_{k})/\|\mathbf{f}_{\mathrm{thumb}}-\mathbf{f}_{k}\|_2$.
The fingertip projections are $f_{k}^{w}=\mathbf{w}_{k}^{\top}\mathbf{f}_{k}$ and $f_{\mathrm{thumb},k}^{w}=\mathbf{w}_{k}^{\top}\mathbf{f}_{\mathrm{thumb}}$.
The object's projected extent is $[p_{\min,k},p_{\max,k}]$, where
$p_{\min,k}=\min_{\mathbf{p}\in\mathcal{P}}\mathbf{w}_{k}^{\top}\mathbf{p}$ and
$p_{\max,k}=\max_{\mathbf{p}\in\mathcal{P}}\mathbf{w}_{k}^{\top}\mathbf{p}$.
GWC measures the fraction of this extent covered by the thumb--finger span:
\begin{equation}
\mathrm{GWC}_{k}=
\frac{\left|[f_{k}^{w},f_{\mathrm{thumb},k}^{w}]
\cap[p_{\min,k},p_{\max,k}]\right|}
{p_{\max,k}-p_{\min,k}}.
\label{eq:gwc}
\end{equation}
Here, $|I|$ denotes the length of a one-dimensional interval, and interval endpoints are sorted in increasing order. A larger GWC indicates greater overlap between the thumb--finger span and the object along the corresponding direction. The measure favors candidates that span the object along the corresponding thumb--finger direction, complementing the depth-based measure below.

\textbf{Grasp Depth Coverage (GDC).}
GDC measures how much of the object's projected depth lies
within the span between a fingertip and the palm reference
plane. It complements GWC by favoring enclosure in the
palm-normal direction rather than relying on lateral
fingertip coverage.

Let $\mathbf{n}_{\mathrm{palm}}$ be the unit palm normal and
$\mathbf{p}_{\mathrm{palm}}$ the palm center.
Define the scalar projections
$d_k=\mathbf{n}_{\mathrm{palm}}^{\top}\mathbf{f}_k$
and
$d_{\mathrm{palm}}=
\mathbf{n}_{\mathrm{palm}}^{\top}\mathbf{p}_{\mathrm{palm}}$.
The object's depth interval is
$[d_{\min},d_{\max}]$, where
$d_{\min}=\min_{\mathbf{p}\in\mathcal{P}}
\mathbf{n}_{\mathrm{palm}}^{\top}\mathbf{p}$
and
$d_{\max}=\max_{\mathbf{p}\in\mathcal{P}}
\mathbf{n}_{\mathrm{palm}}^{\top}\mathbf{p}$.
Then,
\begin{equation}
\mathrm{GDC}_{k}=
\frac{
\left|[d_{\mathrm{palm}},d_k]
\cap[d_{\min},d_{\max}]\right|
}{
d_{\max}-d_{\min}
}
\label{eq:gdc}
\end{equation}
As in GWC, the interval endpoints are sorted in increasing
order before computing the intersection length.
The thumb depth coverage
$\mathrm{GDC}_{\mathrm{thumb}}$
is computed in the same manner using
$d_{\mathrm{thumb}}=
\mathbf{n}_{\mathrm{palm}}^{\top}
\mathbf{f}_{\mathrm{thumb}}$.
Together with GWC, GDC characterizes hand--object
envelopment in complementary directions.

The overall grasp score combines width and depth coverage:
\begin{equation}
\mathrm{Quality}
=
\left(
\sum_{k=1}^{3}
\mathrm{GWC}_{k}\cdot\mathrm{GDC}_{k}
\right)
\mathrm{GDC}_{\mathrm{thumb}}.
\label{eq:quality}
\end{equation}

We select the highest-scoring candidate among those that
survive filtering as the grasp guidance.
The thumb term emphasizes enclosure on the thumb side,
while the combined GWC and GDC terms evaluate complementary
width and depth coverage.
Figure~\ref{fig:Grasp_Guidance_Selection} illustrates the
candidate filtering and ranking process.

\subsection{Policy Learning}
\label{sec:Learning Fast Grasping Policy Using Binary Tactile Sensors}

We train a reinforcement learning policy to coordinate the
mobile base, arm, and dexterous hand using grasp guidance
together with online robot, object, and tactile observations.
Tactile feedback is incorporated into both the policy
observation and the contact-based training reward.

\textbf{Observation.}
The policy observation at time step $t$ is
$\mathbf{O}_{t}
=
(\mathbf{q}_{t},
\mathbf{v}_{t},
\mathbf{o}_{t},
\mathbf{c}_{t},
\mathbf{g})$,
where
$\mathbf{q}_{t}
=
\{
\mathbf{qpos}_{\mathrm{arm},t},
\mathbf{qpos}_{\mathrm{hand},t}
\}$
contains the current arm and hand joint configurations.
All six arm joint positions are observed, although only five
joints are actively controlled.
The mobile-base state
$\mathbf{v}_{t}=\{v_{f,t},v_{y,t}\}$
contains the forward velocity and yaw rate.
The object observation
$\mathbf{o}_{t}
=
\{
\mathbf{pos}_{\mathrm{obj},t},
d_{\mathrm{obj},t}
\}$
contains the object position in the wrist-camera coordinate
system and its distance to the palm center.
The vector $\mathbf{c}_{t}$ contains the binary hand-contact
states obtained from pressure-sensitive tactile sensing.
Finally,
$\mathbf{g}
=
\{
\mathbf{pos}_{g},
\mathbf{rot}_{g},
\mathbf{qpos}_{g}
\}$
denotes the selected grasp guidance, consisting of the grasp
position $\mathbf{pos}_{g}\in\mathbb{R}^{3}$,
orientation $\mathbf{rot}_{g}\in\mathbb{R}^{3}$ represented
by Euler angles, and hand joint configuration
$\mathbf{qpos}_{g}\in\mathbb{R}^{16}$.

\textbf{Action.}
The policy outputs
$\mathbf{A}_{t}
=
\{
\mathbf{v}_{\mathrm{base},t},
\mathbf{qpos}^{\mathrm{cmd}}_{\mathrm{arm},t},
\mathbf{qpos}^{\mathrm{cmd}}_{\mathrm{hand},t}
\}$.
Here,
$\mathbf{v}_{\mathrm{base},t}\in\mathbb{R}^{2}$
contains the commanded forward velocity and yaw rate,
$\mathbf{qpos}^{\mathrm{cmd}}_{\mathrm{arm},t}
\in\mathbb{R}^{5}$
contains the commanded arm joint positions, and
$\mathbf{qpos}^{\mathrm{cmd}}_{\mathrm{hand},t}
\in\mathbb{R}^{16}$
contains the commanded hand joint positions.
The physical arm has six degrees of freedom, while its fourth
joint is fixed during policy execution.

\textbf{Reward Design.}
We design task-specific rewards to encourage
coordinated and rapid mobile grasping while guiding the robot
toward the selected grasp configuration. The rewards regulate
base motion, arm approach, hand configuration, grasp
stabilization, and tactile contact throughout execution. We denote
the end-effector--guidance distance by
$d_g=\|\mathbf{p}_{ee}-\mathbf{p}_{g}\|_2$, and use
$d_g^{xy}$ for its horizontal-plane counterpart.

\textbf{(i) Arm operation radius reward}:
To encourage effective use of the arm workspace,
$R_{\mathrm{radius}}
=
\|\mathbf{p}_{ee}^{xy}-\mathbf{p}_{base}^{xy}\|_2$,
where $\mathbf{p}_{ee}^{xy}$ and
$\mathbf{p}_{base}^{xy}$ denote the horizontal positions of
the end-effector and arm-base reference point, respectively.

\textbf{(ii) Base movement velocity reward}:
To encourage continuous base motion,
$R_{\mathrm{move}}=|v_f|+|v_y|$,
where $v_f$ and $v_y$ denote the forward velocity and yaw
rate, respectively.

\textbf{(iii) Base movement orientation reward}:
To coordinate base motion with the desired approach direction,
$R_{\mathrm{ori}}=-(\omega_b+\omega_v)/\pi$,
where $\omega_b$ and $\omega_v$ are the angular errors between
the desired movement direction and the base heading and
velocity direction, respectively.

\textbf{(iv) Height reward}:
To maintain a suitable approach height,
$R_{\mathrm{height}}=-|z_{ee}-z_a|$
while the object remains on the supporting surface and
$d_g^{xy}>0.5$ m; otherwise it is zero.
Here, $z_a$ denotes the height of the approach point derived
from the selected grasp position and approach direction.

\textbf{(v) Pre-grasp reward}:
Near the target, we align two arm joints with their
guidance-derived target values while additionally reducing
the end-effector position error.
The reward is
$R_{\mathrm{pre}}
=
-\alpha_6 E_6
-
\mathbb{I}(d_g^{xy}\leq d_p)
(\alpha_5E_5+\alpha_pE_p)$,
where
$E_5=
|q^{\mathrm{arm}}_5-q^{g,\mathrm{arm}}_5|/(\pi/2)$,
$E_6=
|q^{\mathrm{arm}}_6-q^{g,\mathrm{arm}}_6|/\pi$,
and $E_p$ is the normalized end-effector position error.

We use
$\alpha_6=1$,
$\alpha_5=2$,
$\alpha_p=0.001$,
and $d_p=0.1$ m.

\textbf{(vi) Hand configuration reward}:
We denote this term by $R_{\mathrm{hand}}$.
The desired hand behavior changes with the distance to the
guided grasp.
When $d_g>0.2$ m, the 16 hand joints are rewarded for moving
toward the open-hand reference
$\mathbf{q}_{\mathrm{open}}=\mathbf{0}\in\mathbb{R}^{16}$;
when $d_g\leq0.2$ m, they are instead rewarded for moving
toward the guided hand configuration
$\mathbf{q}_{g}\in\mathbb{R}^{16}$.
Only joints moving in the desired opening or closing direction
contribute to the reward.
In implementation, the joint-wise alignment scores are
nonlinearly aggregated and clipped to bounded ranges for
stable training.

\textbf{(vii) Rapid arm-motion reward}:
To encourage rapid arm motion during the final approach,
\begin{equation}
R_{\text{reach}}
=
\begin{cases}
\phi(\dot q_1), & d_g\leq0.25,\ \dot q_1>0,\\
0, & \text{otherwise},
\end{cases}
\end{equation}
where $\dot q_1$ denotes the velocity of the first arm joint.
We use $\phi(x)=\exp(x)$ in implementation to assign a larger
reward to faster positive arm motion near the target.

\textbf{(viii) Stable grasping reward}:
To encourage sustained object retention, we define a binary
stable-grasp indicator $c_{\mathrm{stable}}\in\{0,1\}$.
Specifically, $c_{\mathrm{stable}}=1$ when the
end-effector--object distance is below $0.2$ m, its change
between consecutive steps is below $0.01$ m, tactile contact
is detected, and the object has left the supporting region;
otherwise $c_{\mathrm{stable}}=0$.
The reward is
\begin{equation}
R_{\text{stable}}
=
\begin{cases}
\psi(n_t), & c_{\mathrm{stable}}=1,\\
0, & c_{\mathrm{stable}}=0,
\end{cases}
\end{equation}
where $n_t$ denotes the number of consecutive stable-grasp
steps. We use $\psi(n_t)=2^{n_t}$ in implementation to
increasingly reward longer uninterrupted stable grasps.

\textbf{(ix) Tactile reward}:
Beyond the binary contact condition used for grasp
stabilization, we further reward distributed hand--object
contact during stable grasping:
\begin{equation}
R_{\text{tactile}}
=
\begin{cases}
f_{\text{touch}},
& c_{\text{palm}}\neq0
\text{ and } c_{\mathrm{stable}}=1,\\
0, & \text{otherwise},
\end{cases}
\end{equation}
where $c_{\text{palm}}\in\{0,1\}$ denotes the binary
palm-side contact state and
$f_{\text{touch}}=\sum_{i=1}^{N}c_i$
is the number of active finger-contact states.

\textbf{(x) Control penalty}:
To suppress abrupt changes in control commands, we define
$P_{\text{action}}
=\|\mathbf{A}_t-\mathbf{A}_{t-1}\|_2^2$.

The total reward is a weighted combination of these terms:
\begin{equation}
R
=
\sum_i \lambda_i R_i
-
\lambda_a P_{\text{action}},
\end{equation}
where $\lambda_i$ denotes the corresponding reward weight.

\textbf{Policy Optimization.}
We optimize the whole-body grasping policy using Proximal
Policy Optimization (PPO)~\cite{schulman2017proximal},
with advantages estimated using Generalized Advantage
Estimation (GAE).
\begin{table*}[t]
\centering
\caption{\textbf{Simulation results on unseen objects.}
S.R. (\%) and HO.D. (cm) are reported as S.R. (HO.D.).
Each object is evaluated over $10\times500$ grasp trials.
A dash indicates that HO.D. is unavailable because no
successful grasp is obtained.
OS: one-stage; TS: two-stage;
FS: force-direction-based grasp selection adapted
from~\cite{ciocarlie2009hand};
ES: hand-envelopment-based grasp selection.}
\begin{tabular}{l|ccc|ccc}
\toprule
\multirow{2}{*}{Method} &
\multicolumn{3}{c|}{Full Point Cloud} &
\multicolumn{3}{c}{Partial Point Cloud} \\
\cmidrule(lr){2-4} \cmidrule(lr){5-7}
& Easy (60) & Hard (36) & Average
& Easy (60) & Hard (36) & Average \\
\midrule
Reactive Mobile Manipulation~\cite{burgess2023architecture}
& 10.11(7.58) & 5.33(8.11) & 8.31(7.78)
& 10.11(7.58) & 5.33(8.11) & 8.31(7.78) \\

One-Stage (OS)
& 3.13(7.38) & 2.87(7.72) & 3.03(7.51)
& 0(-) & 0(-) & 0(-) \\

TS + FS~\cite{ciocarlie2009hand}
& 52.43(1.95) & 24.60(3.11) & 41.99(2.39)
& 0(-) & 0(-) & 0(-) \\
\midrule

Ours (TS + ES)
& \textbf{59.50(1.62)}
& \textbf{34.42(1.73)}
& \textbf{50.09(1.67)}
& \textbf{47.17(2.67)}
& \textbf{24.09(3.97)}
& \textbf{38.51(3.21)} \\
\bottomrule
\end{tabular}
\label{tab:compare_with_baseline}
\end{table*}
\section{EXPERIMENTS}
\subsection{Experiment Settings}

\begin{figure}[t]
    \centering
    \includegraphics[width=1\linewidth]{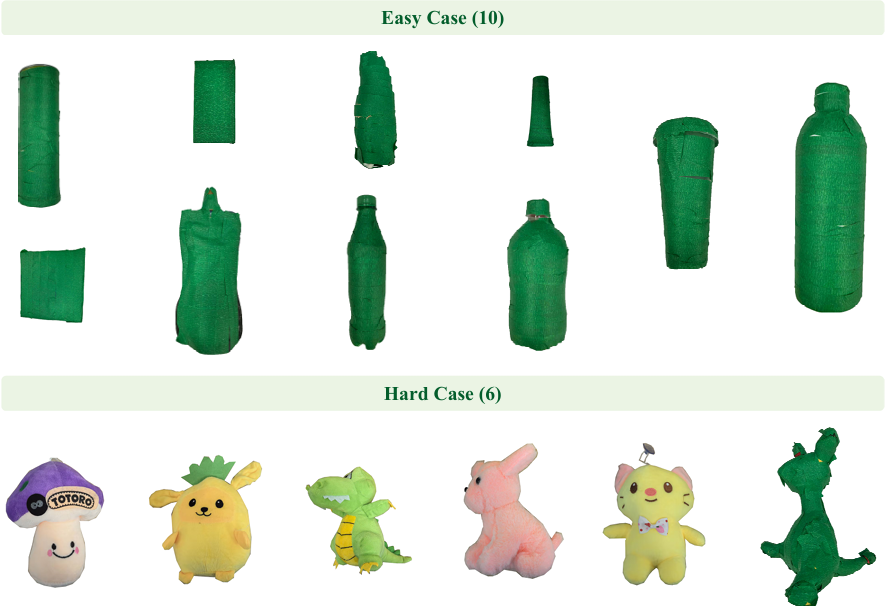}
    \caption{Real-world test objects}
    \label{fig:real obj}
\end{figure}

\textbf{Dataset.}
We use separate datasets for grasp generation and whole-body
policy learning. The grasp synthesis dataset contains 478,200
validated grasps over 4,782 objects, with 100 grasps per
object, generated following~\cite{chen2025bodex}.
For policy learning, we use 418 objects
from~\cite{liu2024realdex,zhong2025dexgrasp}, including
322 training objects and 96 held-out test objects.
The test set contains 60 easy objects with mainly regular,
compact geometries and 36 hard objects with flatter, thinner,
or more irregular shapes.
Real-world evaluation uses 16 objects with varying sizes,
shapes, and weights, comprising 10 easy and 6 hard cases
shown in Fig.~\ref{fig:real obj}.

\textbf{Training and Experimental Setup.}
The grasp generator is trained on the synthetic dataset and
then frozen. For each object, the highest-scoring candidate
that survives filtering is used as grasp guidance.
The policy is trained in 48 parallel environments on an
Intel Core i9-14900K CPU and an NVIDIA GeForce RTX 4090 GPU.
Simulation runs at approximately 60 Hz with policy inference
every four simulation steps, resulting in a 15 Hz policy
frequency, which is also used for physical deployment.

\textbf{Evaluation Metrics.}
We evaluate grasping performance using success rate and
hand--object displacement.

\textbf{(i) Success Rate (S.R.)}:
A trial is considered successful if the object is removed
from the supporting region and remains stably grasped until
the end of the evaluation episode.
Specifically, we use the stable-grasp criterion defined in
Section~\ref{sec:Learning Fast Grasping Policy Using Binary Tactile Sensors},
which requires the end-effector--object distance to be below
$0.2$ m, its change between consecutive steps to be below
$0.01$ m, tactile contact to be maintained, and the object
to remain outside the supporting region.
The success rate is computed as
$\mathrm{S.R.}=N_{\mathrm{success}}/N_{\mathrm{trial}}\times100\%$.

\textbf{(ii) Hand--Object Displacement (HO.D.)}:
HO.D. measures the relative displacement of the object with
respect to the hand after hand--object contact.
A smaller HO.D. indicates less relative object motion during
grasp execution and therefore better object retention.

\begin{table}[t]
\centering
\caption{\textbf{Ablation study.}
Each entry reports S.R. (\%) with HO.D. (cm) in parentheses.
A dash denotes unavailable HO.D. due to zero successful
grasps.}
\label{tab:ablation}
\small
\setlength{\tabcolsep}{3pt}

\begin{tabular}{lccc}
\toprule
Variant & Easy (60) & Hard (36) & Avg. \\
\midrule

\multicolumn{4}{l}{\textit{Reward / penalty}} \\
w/o $R_{\text{radius}}$ & 0(-) & 0(-) & 0(-) \\
w/o $R_{\text{move}}$   & 0(-) & 0(-) & 0(-) \\
w/o $R_{\text{ori}}$    & 0(-) & 0(-) & 0(-) \\
w/o $R_{\text{height}}$ & 32.11(2.86) & 22.67(2.49) & 28.57(2.72) \\
w/o $R_{\text{pre}}$    & 0(-) & 0(-) & 0(-) \\
w/o $R_{\text{hand}}$   & 0(-) & 0(-) & 0(-) \\
w/o $R_{\text{reach}}$  & 0(-) & 0(-) & 0(-) \\
w/o $R_{\text{stable}}$ & 43.51(2.57) & 18.51(2.35) & 34.13(2.49) \\
w/o $P_{\text{action}}$ & 60.83(1.72) & 28.81(1.28) & 48.82(1.56) \\

\midrule
\multicolumn{4}{l}{\textit{Tactile}} \\
w/o $R_{\text{tactile}}$ & 47.80(1.64) & 23.08(1.76) & 38.53(1.69) \\
w/o Tactile obs.         & 43.33(2.62) & 19.64(2.61) & 34.44(2.62) \\

\midrule
\multicolumn{4}{l}{\textit{Grasp guidance}} \\
Random guidance & 0(-) & 0(-) & 0(-) \\
w/o GDC         & 43.52(1.67) & 19.64(2.61) & 34.57(2.02) \\
w/o GWC         & 18.90(4.99) & 7.04(3.76) & 14.45(4.53) \\

\midrule
Ours & 59.50(1.62) & 34.42(1.73) & 50.09(1.67) \\
\bottomrule
\end{tabular}
\end{table}

\subsection{Results on Fast Grasping}

We compare FastGrasp with three baselines in simulation
under both full and partial point-cloud observations:
(i) the reactive mobile manipulation method
of~\cite{burgess2023architecture};
(ii) a one-stage baseline (OS) that directly uses PointNet
features~\cite{qi2017pointnet} without explicit grasp
guidance; and
(iii) a two-stage baseline using force-direction-based grasp
selection (TS+FS), where the selection criterion is adapted
from~\cite{ciocarlie2009hand}.
TS+FS uses the same generated grasp candidates as FastGrasp
but replaces our envelopment-based selection criterion with
the force-direction-based quality metric.

As shown in Table~\ref{tab:compare_with_baseline},
FastGrasp achieves the highest S.R. and lowest HO.D. across
all evaluated settings.
Under full point-cloud observations, FastGrasp achieves an
average S.R. of $50.09\%$, compared with $41.99\%$ for
TS+FS, corresponding to an improvement of $8.10$ percentage
points.
The improvement is more pronounced on hard objects, where
S.R. increases from $24.60\%$ to $34.42\%$.
FastGrasp also reduces the average HO.D. from $2.39$ cm for
TS+FS to $1.67$ cm, indicating less relative object motion
during grasp execution.

Under partial point-cloud observations, FastGrasp retains an
average S.R. of $38.51\%$, whereas OS and TS+FS obtain no
successful grasps under the same setting.
The reactive mobile manipulation baseline achieves an average
S.R. of $8.31\%$ and an HO.D. of $7.78$ cm, while FastGrasp
achieves an HO.D. of $3.21$ cm.
In our implementation of the reactive baseline, the target
object is represented by its coordinate rather than by a point
cloud, and therefore its results remain unchanged between the
full and partial point-cloud settings.
Overall, these results show that the proposed
envelopment-based grasp guidance improves grasp success
and reduces relative object motion during grasp execution,
particularly when only partial object geometry is available.
\begin{table*}[t]
\centering
\caption{\textbf{Real-world S.R. (\%).}
Each object is evaluated in 10 consecutive trials.}
\begin{tabular}{l|ccc|ccc}
\toprule
\multirow{2}{*}{Method}
& \multicolumn{3}{c|}{
$v_f^{\max}=1.3\,\mathrm{m/s}$,
$v_y^{\max}=1.0\,\mathrm{rad/s}$}
& \multicolumn{3}{c}{
$v_f^{\max}=0.65\,\mathrm{m/s}$,
$v_y^{\max}=0.5\,\mathrm{rad/s}$} \\
\cmidrule(lr){2-4} \cmidrule(lr){5-7}
& Easy (10) & Hard (6) & Avg.
& Easy (10) & Hard (6) & Avg. \\
\midrule
Ours w/o DR
& 0 & 0 & 0
& 0 & 0 & 0 \\

Ours w/o LPF
& 0 & 0 & 0
& 0 & 0 & 0 \\

Ours w/o TA
& 31 & 20 & 26.9
& 33 & 24 & 29.6 \\
\midrule

\textbf{Ours}
& \textbf{35}
& \textbf{27}
& \textbf{32.0}
& \textbf{38}
& \textbf{29}
& \textbf{34.6} \\
\bottomrule
\end{tabular}
\label{tab:realworld}
\end{table*}

\begin{figure}[!t]
    \centering
    \includegraphics[width=1\linewidth]{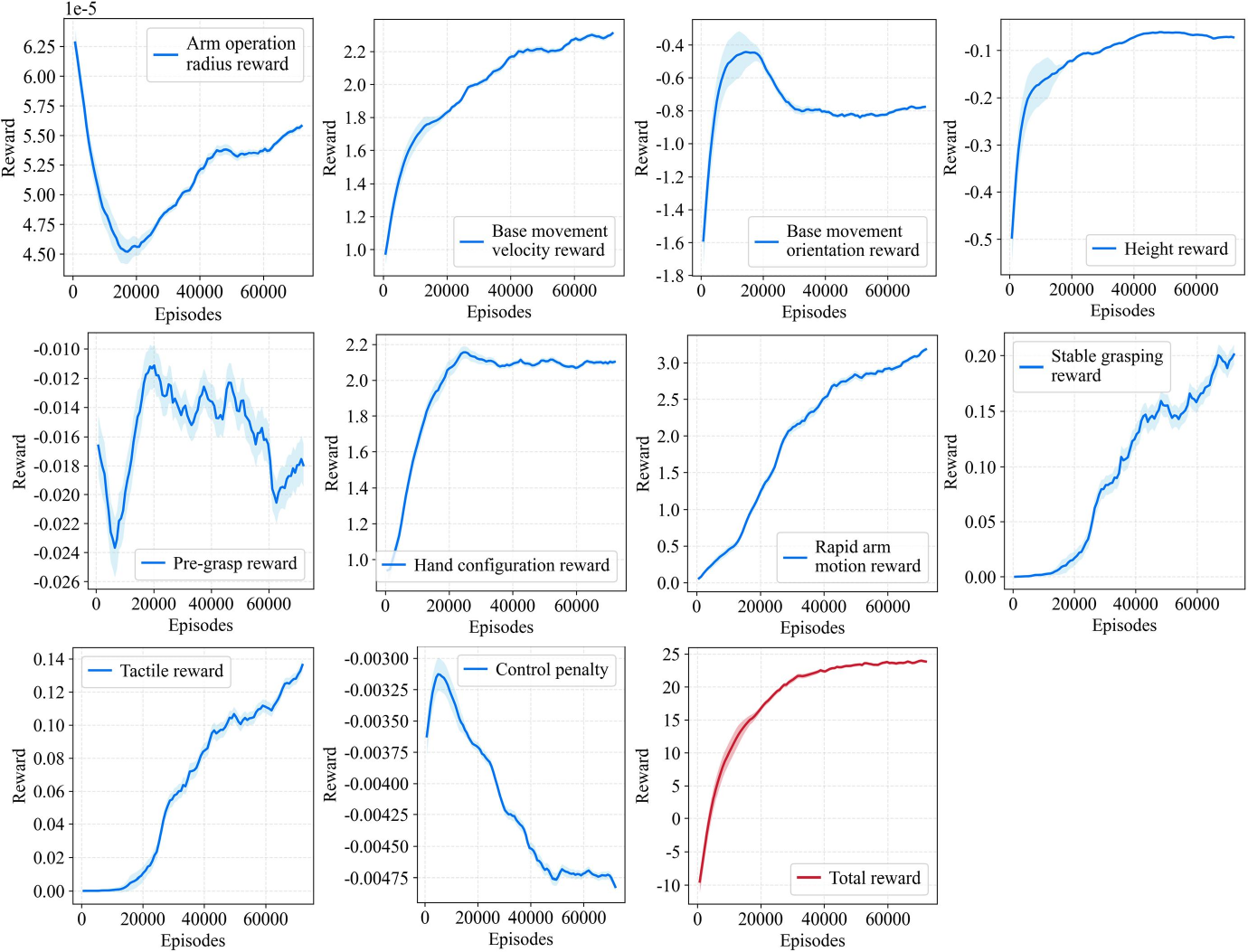}
    \caption{Training reward curves over episodes.
    The red curve denotes the total reward, while the blue
    curves show reward components associated with different
    phases of grasp execution.}
    \label{fig:reward_curve}
\end{figure}

Fig.~\ref{fig:reward_curve} shows the evolution of reward
components during training.
Approach-related rewards increase as the policy learns
coordinated base--arm motion, while pre-grasp and
hand-configuration rewards stabilize as the robot aligns with
the grasp guidance.
Stable-grasp and tactile rewards become increasingly active
during grasp execution.
These trends are consistent with the intended progression
from approach to grasp stabilization, while individual reward
contributions are evaluated in Table~\ref{tab:ablation}.

\subsection{Ablation Study}

Table~\ref{tab:ablation} evaluates the reward and tactile
components.
Removing $R_{\mathrm{radius}}$, $R_{\mathrm{move}}$,
$R_{\mathrm{ori}}$, $R_{\mathrm{pre}}$,
$R_{\mathrm{hand}}$, or $R_{\mathrm{reach}}$ results in no
successful grasps, while removing $R_{\mathrm{height}}$ or
$R_{\mathrm{stable}}$ reduces the average S.R. to $28.57\%$
and $34.13\%$, respectively.
Removing the action-rate penalty yields a similar S.R.
($48.82\%$ versus $50.09\%$), consistent with its primary
role as a control-smoothness regularizer.
Removing tactile observations and the tactile reward reduces
S.R. by $15.65$ and $11.56$ percentage points, respectively;
without tactile observations, HO.D. increases from
$1.67$ cm to $2.62$ cm.

For grasp-guidance selection, random candidate selection
produces no successful grasps.
Removing GDC reduces the average S.R. to $34.57\%$, while
removing GWC reduces it to $14.45\%$ and increases HO.D. to
$4.53$ cm.
Thus, both coverage terms contribute to effective grasp
guidance, with GWC having a stronger effect in the evaluated
setting.

\subsection{Real Robot Deployment}

\textbf{Sim2Real Transfer.}
We use four components to reduce sensing and actuation
discrepancies between simulation and the physical system.
\textbf{(i) Object observation:}
A simulated RGB-D camera matching the real setup provides
point-cloud observations, reducing the gap between simulated
state information and real-world visual sensing.
\textbf{(ii) Low-Pass Filtering (LPF):}
To accommodate actuator dynamics and suppress high-frequency
command variations, we smooth the base velocity and arm/hand
position commands:
\begin{equation}
c_{\mathrm{filtered}}[t]
=
\alpha c_{\mathrm{filtered}}[t-1]
+
(1-\alpha)c_{\mathrm{raw}}[t],
\end{equation}
where $\alpha=0.3$, and $c_{\mathrm{raw}}$ and
$c_{\mathrm{filtered}}$ denote the raw and filtered commands,
respectively.
\textbf{(iii) Domain Randomization (DR):}
We randomize the environment settings described in
Section~\ref{sec:System_Setup}, inject observation and
action noise, and perturb joint-related parameters and
friction to improve robustness to sensing and actuation
uncertainties~\cite{tobin2017domain}.
\textbf{(iv) Tactile Adaptation (TA):}
During physical deployment, binary pressure feedback triggers
finger tightening to support timely grasp closure after contact.
TA complements the learned whole-body controller and is
distinct from the tactile observations and tactile reward
used during policy learning.

\textbf{Real-world Results.}
As shown in Table~\ref{tab:realworld}, FastGrasp achieves
average S.R.s of $32.0\%$ under the primary high-speed
setting and $34.6\%$ under the reduced-speed reference
setting, compared with $26.9\%$ and $29.6\%$ without TA.
Removing either DR or LPF results in no successful grasps,
highlighting their importance for physical deployment.
The slightly higher S.R. under the reduced-speed setting is
consistent with its less aggressive overall motion
constraints, since both forward-velocity and yaw-rate limits
are reduced.
TA improves but is not required for all successful grasps,
indicating that tactile-triggered tightening complements the
learned whole-body controller.

\textbf{Safety.}
During simulation training, an unintended collision
terminates the current episode.
During real-world experiments, a human operator continuously
monitors each trial and manually stops the system if unsafe
motion or a potential collision is observed.

\section{Conclusion}

We presented FastGrasp, a two-stage learning framework for
fast dexterous grasping under continuous mobile-base motion.
FastGrasp combines grasp-guidance selection, whole-body
reinforcement learning, and tactile feedback.
Grasp candidates are filtered for approach executability and
supporting-surface compatibility and ranked by hand--object
envelopment, while the selected grasp guides coordinated
base--arm--hand control.
Binary tactile feedback further supports contact-aware policy
learning and deployment-time grasp tightening.
Experiments on held-out objects demonstrate higher grasp
success and lower hand--object displacement under both full
and partial observations, and real-world results demonstrate
transfer across diverse object geometries.

The current system remains challenged by flat or highly
irregular objects and by safe high-speed operation in complex
environments.
Future work will explore more adaptive grasp guidance,
safer whole-body control, and tighter integration of fast
navigation with dexterous manipulation.

 \subsubsection*{Acknowledgments}
This work was supported by MoE Key Laboratory of Intelligent Perception and Human-Machine Collaboration (KLIP-HuMaCo), Shanghai Frontiers Science Center of Human-centered Artificial Intelligence (ShangHAI).

\bibliographystyle{IEEEtran}
\bibliography{iros_references}

\end{document}